\documentclass[runningheads]{llncs}

\usepackage[T1]{fontenc}
\usepackage{graphicx,verbatim}
\usepackage{amsmath,amssymb}
\usepackage{booktabs,multirow}
\usepackage{pifont}
\usepackage{url}
\usepackage[hidelinks]{hyperref}
\usepackage[misc]{ifsym}
\usepackage{microtype}

\newcommand{\cmark}{\ding{51}}

\begin{document}

\title{Test-Time Adaptation for ECG Classification via SQI-Gated Self-Training and Beat-Rhythm Consistency}

\author{Wenhan Jiang\inst{1} \and
Zhipeng Deng\inst{1} \textsuperscript{\Letter} \and
Jiale Zhou\inst{1} \and
Haolin Wang\inst{2} \and
Yafei Ou\inst{3} \and
Yefeng Zheng\inst{1} \textsuperscript{\Letter}}

\authorrunning{W. Jiang et al.}

\institute{
Westlake University, Hangzhou, China\\
\and
Hokkaido University, Sapporo, Japan\\
\and
RIKEN, Japan\\
\email{
\{jiangwenhan, dengzhipeng, zhoujiale, zhengyefeng\}@westlake.edu.cn;
yafei.ou@riken.jp
}
}

\titlerunning{BeatRhythm-TTA}

\maketitle

\begin{abstract}
Deep learning models for electrocardiogram (ECG) classification often suffer from significant performance degradation when deployed in unseen domains due to shifts in acquisition devices and patient populations. Test-time adaptation (TTA) offers a practical solution by adapting models using only unlabeled data at inference time. However, existing TTA methods often underperform on ECG tasks, since naive online updates ignore the hierarchical beat-rhythm structure of cardiac cycles and are vulnerable to signal artifacts, which leads to unstable adaptation and model drift. We propose \textbf{BeatRhythm-TTA}, an ECG-tailored TTA framework that explicitly accounts for ECG's noisy observations and structured beat-rhythm semantics under domain shift. First, to handle pervasive ECG artifacts, we introduce a Signal Quality Index (SQI)-gated adaptation scheme that selectively filters out low-quality signals to prevent harmful updates. Second, to leverage ECG's beat-rhythm semantics, we enforce dual-level consistency so the model preserves beat morphology and rhythm dynamics while adapting to shifted acquisition conditions. Extensive experiments on multi-label ECG diagnosis across three adaptation protocols, using PTB-XL as the source domain and CPSC2018/Georgia as two target domains, demonstrate the effectiveness of our method, yielding an average +2.70\% relative improvement in Macro-F1 over the best competing method. The code is available at \url{https://github.com/Vincent-Wenhan/BeatRhythm-TTA}.

\end{abstract}

\section{Introduction}

\begin{figure}
    \centering
    \includegraphics[width=\linewidth]{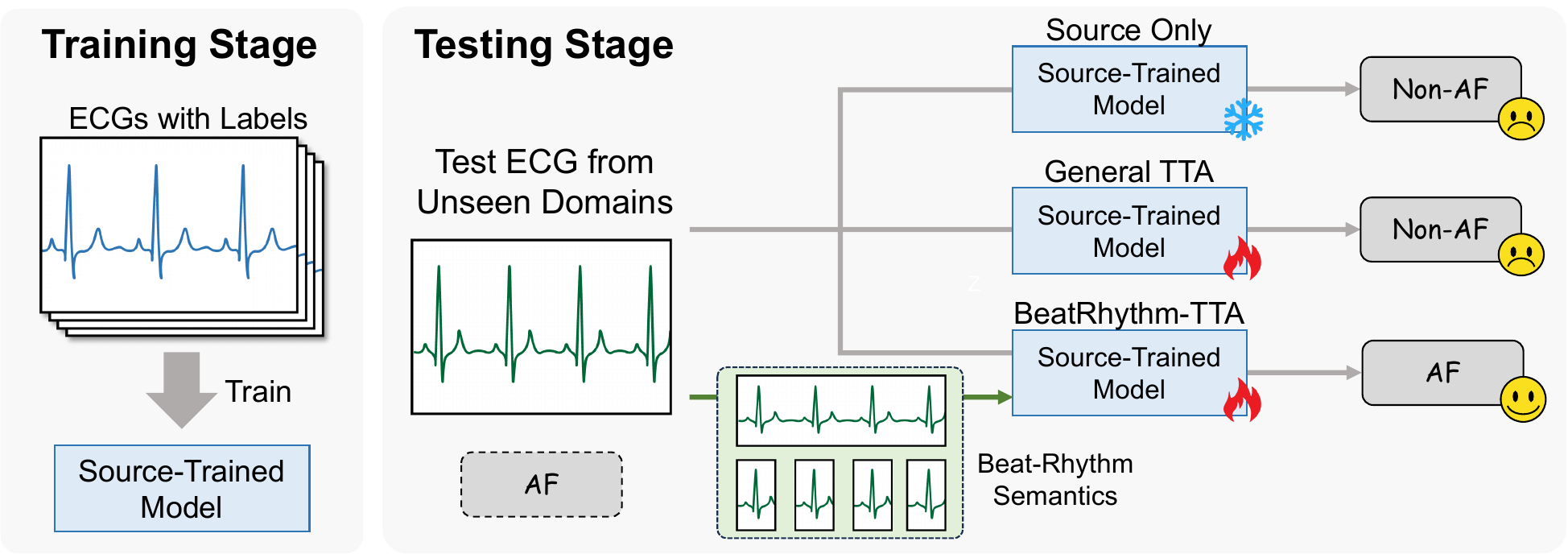}
    \caption{\textbf{Motivation of the proposed BeatRhythm-TTA.}
Comparison of TTA strategies. Generic TTA methods are brittle on ECG, while BeatRhythm-TTA leverages beat-rhythm semantics for robust adaptation. (AF denotes atrial fibrillation, a cardiac rhythm disorder.)}
    \label{fig:motivation}
\end{figure}

Although deep learning has achieved strong performance on electrocardiogram (ECG) interpretation~\cite{liu2023spectral,ribeiro2020automatic}, generalizing pretrained models to unseen clinical environments remains challenging due to distribution shifts introduced by heterogeneous hardware and patient cohorts~\cite{huang2024generalization,ong2024shortcut}.
Re-training or fine-tuning for each new target domain is often impractical in clinical practice as it requires substantial expert annotation and repeated engineering effort~\cite{ong2024shortcut}.
This motivates adopting test-time adaptation (TTA) from the broader machine learning literature~\cite{wang2020tent,wang2022continual}, a paradigm that adapts a deployed model using only unlabeled test data, without requiring target labels or retraining from scratch.

Most TTA methods have been developed and validated primarily on vision benchmarks~\cite{wang2020tent,wang2022continual}. These methods are usually based on normalization calibration (e.g., AdaBN~\cite{li2016revisiting}, MemBN~\cite{kang2024membn}), entropy minimization (e.g., TENT~\cite{wang2020tent}, SAR~\cite{niu2023towards}, DeYO~\cite{lee2024entropy}), or confidence- or energy-based objectives (e.g., COME~\cite{zhang2024come}, TEA~\cite{yuan2024tea}).
Recent work has extended TTA beyond vision, showing promising performance in speech, audio, and AI-for-science under distribution shifts~\cite{kim2023sgem,kim2025battling,zhang2024pass,zhou2025topotta}.
A closely related paradigm is source-free domain adaptation (SFDA)~\cite{hwang2024sf,liang2020we,xu2025revisiting}, which similarly addresses domain shifts without source data but is typically formulated as an offline adaptation stage before deployment.
Despite progress in other modalities, ECG-specific TTA remains underexplored, as directly transferring existing methods raises modality-specific challenges.
First, adapting models on continuous ECG streams corrupted by transient artifacts (e.g., baseline wander, motion artifacts) amplifies spurious correlations and triggers drift. Second, many existing TTA objectives (e.g., global entropy minimization) ignore the hierarchical beat-rhythm semantics that underlie ECG diagnosis~\cite{carrington2022monitoring,vogel2019st}.

To this end, we propose BeatRhythm-TTA, a semantically grounded TTA framework comprising two complementary components. First, \textbf{SQI-gated adaptation} selectively scales or blocks test-time updates based on a signal quality index (SQI) to avoid harmful adaptation on noisy segments. Second, \textbf{dual-level consistency} enforces beat- and rhythm-level agreement to preserve physiologically motivated semantics at two temporal scales~\cite{hong2019mina}: local beat morphology (e.g., ST-segment changes~\cite{vogel2019st}) and global rhythm irregularities (e.g., atrial fibrillation~\cite{carrington2022monitoring}).
By coupling reliability-aware gating with physiologically grounded consistency, BeatRhythm-TTA reduces drift under real-world domain shifts. Extensive experiments on multi-label ECG classification across three adaptation protocols, using PTB-XL as the source domain and CPSC2018/Georgia as two target domains, validated the effectiveness of our approach. It outperforms the strongest competing methods with an average +2.70\% relative improvement in Macro-F1 across protocols.

\section{Method}

\begin{figure}[t!]
    \centering
    \includegraphics[width=\linewidth]{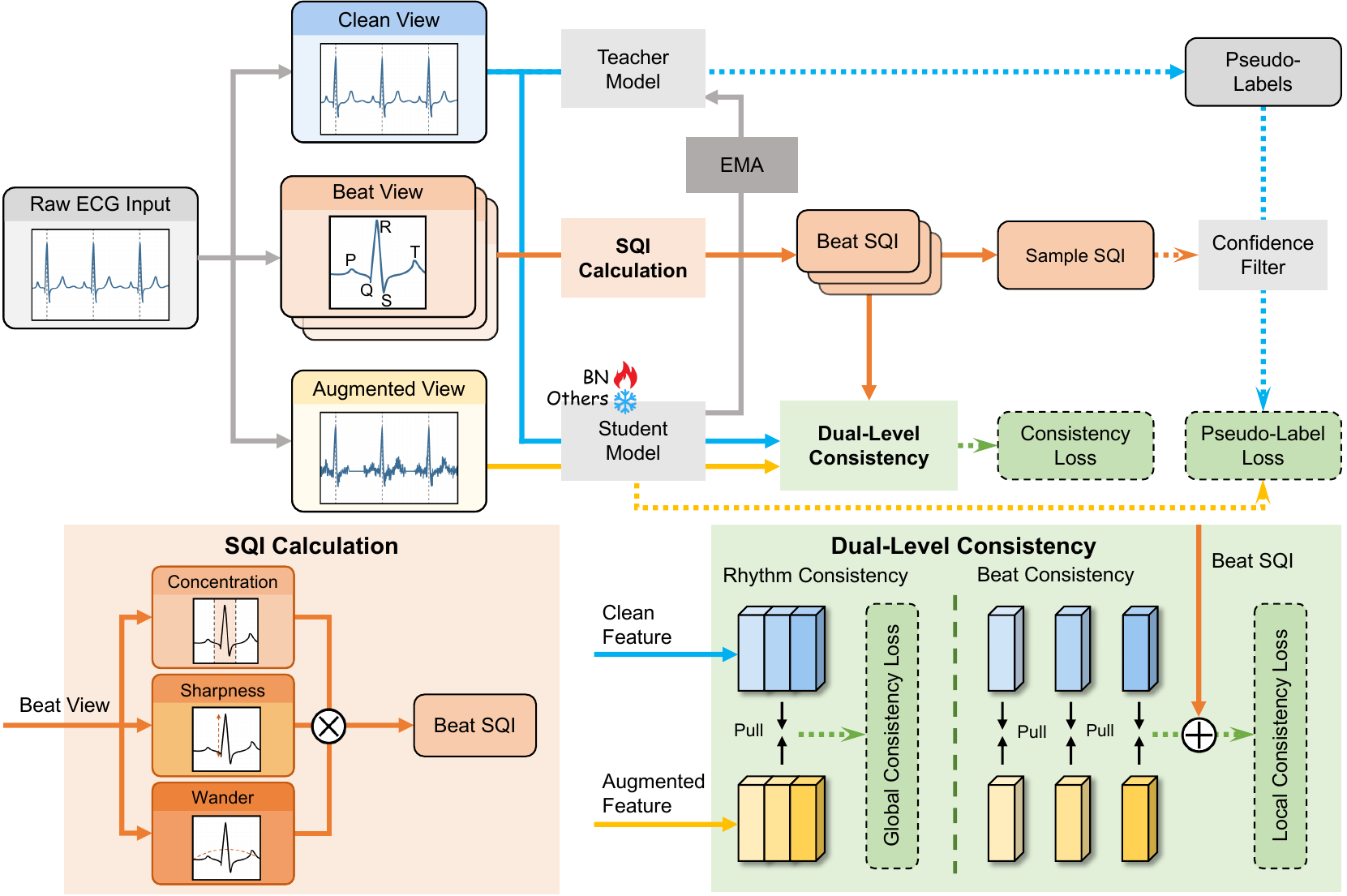}
    \caption{\textbf{BeatRhythm-TTA Framework.} Given an unlabeled target ECG, adaptation is driven by two complementary modules. \textbf{(1) SQI-gated pseudo-labeling:} Blocks self-training updates based on teacher confidence and beat-wise signal quality, where EMA (exponential moving average) maintains a slowly updated teacher for more stable pseudo-labels. \textbf{(2) Dual-level consistency:} Enforces rhythm- and beat-level agreement between clean and augmented views to stabilize adaptation.}
    \label{fig:method_overview}
\end{figure}

\subsection{Preliminaries and Overview}

\textbf{Problem definition.}
Let $f(\cdot;\theta)$ be an ECG classifier that maps a $C$-lead segment $x\in\mathbb{R}^{C\times T}$ to logits $z\in\mathbb{R}^{K}$, with multi-label targets $y\in\{0,1\}^{K}$, where $K$ denotes the number of diseases to classify.
We first train $\theta^\mathcal{S}$ on the labeled source domain $\mathcal{D}^\mathcal{S}=\{(x_i^\mathcal{S},y_i^\mathcal{S})\}_{i=1}^{N^\mathcal{S}}$:
\begin{equation}
\theta^\mathcal{S}=\arg\min_{\theta}\;\mathbb{E}_{(x^\mathcal{S},y^\mathcal{S})\sim\mathcal{D}^\mathcal{S}}\,\mathcal{L}_{\text{sup}}\!\left(f(x^\mathcal{S};\theta),y^\mathcal{S}\right).
\end{equation}
At deployment, unlabeled target samples $\{x_i^\mathcal{T}\}_{i=1}^{N^\mathcal{T}}$ arrive from an unseen domain $\mathcal{D}^\mathcal{T}$.
Test-time adaptation (TTA) updates the model using only $x^\mathcal{T}$ by minimizing a self-supervised loss:
\begin{equation}
\theta^{\mathcal{T}}=\arg\min_{\theta}\;\mathbb{E}_{x^{\mathcal{T}}\sim\mathcal{D}^{\mathcal{T}}}\,
\mathcal{L}_{\text{tta}}\!\left(f(x^{\mathcal{T}};\theta),x^{\mathcal{T}}\right).
\end{equation}

\textbf{Offline and online TTA settings.}
We consider both offline and online test-time adaptation settings.
Offline TTA assumes the entire target test distribution $\{x_i^\mathcal{T}\}_{i=1}^{N^\mathcal{T}}$ is available for adaptation and we optimize $\mathcal{L}_{\text{tta}}$ on the unlabeled target test set and then evaluate on the same set.
Online TTA processes test samples sequentially: as each sample arrives, the model is updated on it and predicts on-the-fly.
We consider three protocols: (i) offline, (ii) continual online, and (iii) independent online TTA. Continual online carries updates across samples ($\theta_i \rightarrow \theta_{i+1}$), while independent online resets to $\theta^{\mathcal{S}}$ for each sample.

\textbf{Method Overview.}
The overview of our method is shown in Fig.~\ref{fig:method_overview}.
In real ECG recordings, signal quality can vary due to artifacts, and diagnostic semantics appear at both the beat and rhythm levels.
Motivated by these properties, we propose BeatRhythm-TTA, an ECG-tailored framework with two complementary modules: an SQI-gated pseudo-labeling module that computes a beat-wise SQI to gate teacher-driven self-training, and a dual-level consistency module that enforces rhythm- and beat-level agreement across clean and augmented views.

\subsection{SQI-gated Pseudo-labeling}

To prevent harmful updates triggered by noisy samples and segments, we perform selective self-training at test time, only updating the model when the current ECG sample is both confident under a teacher model and reliable according to a signal quality index (SQI).
Signal quality indices have been widely used to quantify waveform reliability in ECG analysis~\cite{clifford2012signal,rahman2022robustness}. Unlike prior SQIs that mainly target record-level screening and may require heavier, domain-tuned features, our SQI is a lightweight beat-wise score designed for test-time gating, computed online without labels to control update eligibility and strength.

Given an unlabeled ECG sample $x\!\in\!\mathbb{R}^{C\times T}$, we detect R-peaks $\{r_i\}_{i=1}^{M_b}$ using the WFDB toolchain~\cite{goldberger2000components}, where $M_b$ denotes the number of detected beats in the sample. We select a reference lead based on derivative energy and apply robust normalization with basic physiological sanity checks.
We then extract beat-centered windows and scale the window length by the median RR interval to account for heart-rate variation.
For each beat window $x^{(i)}$, we compute three lightweight morphology factors:

\textbf{(i) QRS concentration $(\mathrm{conc}^{(i)})$}: the QRS complex, the sharp waveform around the R-peak in the P-QRS-T beat structure, carries concentrated energy in clean beats~\cite{wagner2013marriott};
we measure it as the energy ratio of a small center region to the whole beat window;

\textbf{(ii) sharpness $(\mathrm{sharp}^{(i)})$}: true QRS has steep slopes, measured by a high-quantile magnitude of the first-order difference normalized by signal root mean square;

\textbf{(iii) baseline wander ratio $(\mathrm{bwr}^{(i)})$}: strong low-frequency drift, measured as the energy ratio of a moving-average baseline estimate to the whole window, indicates poor motion.

We combine them to obtain a beat SQI score $\mathrm{SQI}^{(i)}$:
\begin{equation}
\mathrm{SQI}^{(i)} \;=\;
\sigma\!(\mathrm{conc}^{(i)})\cdot
\sigma\!(\mathrm{sharp}^{(i)})\cdot
\sigma\!(-\mathrm{bwr}^{(i)}),
\label{eq:sqi}
\end{equation}
where $\sigma(\cdot)$ is the sigmoid function.
Finally, we aggregate beat-wise scores into a sample-level quality weight $w=\frac{1}{M_b}\sum_i \mathrm{SQI}^{(i)}$ (mean over beats).

We adopt a teacher-student scheme to generate pseudo labels from a clean view and supervise the student on an augmented view as:
\begin{equation}
\mathcal{L}_{\mathrm{pl}}
=
\mathbb{I}\!\left[c(x)\ge\tau_c\right]\, \mathbb{I}\!\left[w(x)\ge\tau_q\right]\;
\ell_{bce}\!\Big(p_{\theta}(\tilde{x}),\; p_{\bar{\theta}}(x^{c})\Big),
\label{eq:pl_sqi}
\end{equation}
where $x^{c}=x$ is the clean view, $\tilde{x}=\mathcal{A}(x)$ is a time-aligned augmentation, and $\ell_{bce}$ denotes binary cross-entropy (BCE) loss.
We denote $p_{\theta}(x)=\sigma\!\left(f_{\theta}(x)\right)$ as multi-label probabilities, and use a teacher $f_{\bar{\theta}}$ to produce soft pseudo-labels $p_{\bar{\theta}}(x^{c})$.
We define the confidence score as $c(x)=\min_{k}\left|\,p_{\bar{\theta}}(x^{c})_{k}-\frac{1}{2}\right|$, and $\mathbb{I}[\cdot]$ denotes the indicator function, so activate pseudo-label updates only when both the confidence and the SQI thresholds are satisfied.
The teacher is updated by exponential moving average (EMA)~\cite{tarvainen2017mean,wang2022continual}:
$\bar{\theta}\leftarrow \mu\bar{\theta}+(1-\mu)\theta$.
This EMA update stabilizes pseudo-labels and mitigates rapid forgetting during online adaptation.

\subsection{Dual-level Consistency}

ECG semantics appear at two temporal scales: beat-level morphology and rhythm-level dynamics across beats~\cite{hong2019mina}. Diagnoses may rely on subtle beat morphology such as ST-segment elevation/depression and T-wave inversion~\cite{vogel2019st} or rhythm irregularity across multiple beats such as atrial fibrillation (AF)~\cite{carrington2022monitoring}. Accordingly, to stabilize adaptation while preserving clinically meaningful morphology, we regularize test-time updates with a dual-level consistency objective that aligns global sample representations and local beat-centric representations between a clean view $x^{c}$ and a time-aligned augmented view $\tilde{x}$.

Let $F(\cdot)$ denote intermediate features extracted from a chosen backbone layer.
Given the two time-aligned views, we obtain feature of the clean view ($F_c=F(x^{c})$), the augmented view ($F_a=F(\tilde{x})$), and minimize a cosine alignment loss:
\begin{equation}
\mathcal{L}_{\mathrm{rhythm}}
= 1 - \cos\!\left(\mathrm{sg}(F_c),\, F_a\right),
\label{eq:l_global}
\end{equation}
where $\mathrm{sg}(\cdot)$ denotes stop-gradient operator.

Rhythm alignment captures global dynamics but does not explicitly constrain beat morphology.
Using the previously detected R-peaks $\{r_i\}$ and the RR-based windowing, we extract beat-centric embeddings from $F_c$ and $F_a$, yielding paired sets $\{b_c^{(i)}\}_{i=1}^{M_b}$ and $\{b_a^{(i)}\}_{i=1}^{M_b}$.
We align beat embeddings using cosine distance:
\begin{equation}
\ell^{(i)}_{\mathrm{beat}} = 1 - \cos\!\left(\mathrm{sg}(b^{(i)}_c),\, b^{(i)}_a\right),
\label{eq:l_beat_vec}
\end{equation}
and aggregate across beats using the beat-wise SQI weights from Eq.~(\ref{eq:sqi}), emphasizing reliable beats and suppressing corrupted ones during adaptation:
\begin{equation}
\mathcal{L}_{\mathrm{beat}} =
\frac{\sum_{i}\,\omega^{(i)}\,\ell^{(i)}_{\mathrm{beat}}}{\sum_{i}\,\omega^{(i)}+\epsilon},
\qquad
\omega^{(i)}=\max(\mathrm{SQI}^{(i)},\tau_{\mathrm{floor}}),
\label{eq:l_beat}
\end{equation}
where $\tau_{\mathrm{floor}} = 0.05$ sets a minimum beat weight.

We combine the dual-level consistency losses with the SQI-gated pseudo-label loss:
\begin{equation}
\mathcal{L}=\mathcal{L}_{\mathrm{pl}}+\lambda_{\mathrm{beat}}\mathcal{L}_{\mathrm{beat}}+\lambda_{\mathrm{rhythm}}\mathcal{L}_{\mathrm{rhythm}}.
\label{eq:overall}
\end{equation}
$\lambda_{\mathrm{beat}}$ and $\lambda_{\mathrm{rhythm}}$ are scalar hyperparameters that balance the contributions of beat-level and rhythm-level consistency against the pseudo-labeling loss.
By optimizing this objective, the model is updated via SQI-gated self-training, while dual-level consistency regularizes test-time updates to preserve both beat- and rhythm-level semantics.

\section{Experiments}

\subsection{Datasets and Experiment Setup}

We evaluated our method on three public 12-lead ECG datasets from the PhysioNet/CinC Challenge 2020~\cite{perez2020classification}. Specifically, we used PTB-XL (Germany, 21,837 records) as the source domain, and evaluated cross-domain adaptation on CPSC2018 (China, 6,877 records) and Georgia (USA, 10,344 records) as two distinct target domains. All datasets were projected into a shared 6-class label space (atrial fibrillation, first-degree atrioventricular block, left bundle branch block, right bundle branch block, premature atrial contraction, and normal sinus rhythm), discarding unshared labels and empty samples.

\textbf{Evaluation metrics.}
Macro-F1 (the unweighted mean of per-class F1 scores) and Macro-AUC (the unweighted mean of per-class AUROC) were reported on the target test sets after adaptation.
Macro averaging gives equal weight to each class, which is important under class imbalance in multi-label ECG diagnosis. All the results were averaged over 3 runs.

\textbf{Implementation details.}
Experiments were implemented in PyTorch on an RTX 4090 GPU.
For preprocessing, all ECGs were resampled to 500\,Hz and padded to a fixed length of 10\,s.
Following~\cite{abbaspourazad2023large,gao2025echoingecg,liu2024zero}, a 1D ResNet-18 was pretrained on PTB-XL (8:2 train/val split) for 100 epochs using the BCE loss, standard augmentations, and the Adam optimizer ($lr\!=\!10^{-4}$, batch size 512). The best validation checkpoint was selected as the source model to initialize TTA. Offline TTA adapted on the target set for 1 epoch (batch size 256). Continual and independent online TTA processed single samples (batch size 1) for 5 and 10 internal steps using the Adam optimizer with $lr\!=\!10^{-6}$ and $10^{-4}$, respectively. For our method, we disabled dropout and updated only BatchNorm affine parameters. Thresholds in Eq.~(\ref{eq:pl_sqi}) were set to $\tau_c\!=\!0.2$ and $\tau_q\!=\!0.05$. The loss weights in Eq.~(\ref{eq:overall}) were set to $\lambda_{\mathrm{beat}}\!=\!0.5$ and $\lambda_{\mathrm{rhythm}}\!=\!1.0$ for all experiments.

\subsection{Results and Analysis}

We compared our method against the source-only baseline and representative TTA methods (AdaBN~\cite{li2016revisiting}, MemBN~\cite{kang2024membn}, TENT~\cite{wang2020tent}, SAR~\cite{niu2023towards}, DeYO~\cite{lee2024entropy}, COME~\cite{zhang2024come} and TEA~\cite{yuan2024tea}). For the offline setting, we additionally evaluated source-free domain adaptation baselines (SHOT~\cite{liang2020we}, SF(DA)$^2$~\cite{hwang2024sf} and UCon~\cite{xu2025revisiting}).
Tables~\ref{tab:offline_tta}--\ref{tab:online_tta} show that our method consistently achieves the best average Macro-AUC and Macro-F1 across offline and online protocols.

\textbf{Offline TTA.}
Under batch adaptation (Table~\ref{tab:offline_tta}), our method improves the averaged Macro-F1 from 60.64\% to 62.42\%, and Macro-AUC from 88.13\% to 88.84\%.
This suggests that combining SQI-gated updates with dual-level consistency provides reliable optimization signals when the target set can be revisited.

\textbf{Continual online TTA.}
In the streaming setting without resets (Table~\ref{tab:online_tta}), updates accumulate and the model can be influenced by occasional low-quality segments.
Our method remains robust and yields the best averaged metrics, improving Macro-F1 by +1.12\% and Macro-AUC by +0.65\%.
This highlights the benefit of reliability-aware SQI gating for preventing unstable updates in long test streams under continual online TTA setting.

\textbf{Independent online TTA.}
Also shown in Table~\ref{tab:online_tta}, our method achieves the highest average performance in the independent setting, improving Macro-F1 by +1.97\%. ECG samples vary widely across individuals in heart rate, morphology, and noise patterns. By combining SQI-gated updates with beat-rhythm consistency, our method adapts to each sample in a personalized and stable manner, yielding the largest gains.

\begin{table}[!t]
\centering
\caption{Offline TTA results. \textbf{Bold}: best, \underline{underlined}: second best. All values are reported in percent (\%).}
\label{tab:offline_tta}
\setlength{\tabcolsep}{5pt}
\begin{tabular}{l|cc|cc|cc}
\toprule
\textbf{Method} &
\multicolumn{2}{c|}{\textbf{CPSC2018}} &
\multicolumn{2}{c|}{\textbf{Georgia}} &
\multicolumn{2}{c}{\textbf{Avg}} \\
& \textbf{AUC} & \textbf{F1} & \textbf{AUC} & \textbf{F1} & \textbf{AUC} & \textbf{F1} \\
\midrule
\textbf{Source Only} & 87.62 & 60.84 & 88.59 & 60.35 & 88.11 & 60.59 \\
\midrule
AdaBN~\cite{li2016revisiting}        & 83.25 & 46.71 & 87.33 & 52.29 & 85.29 & 49.50 \\
TENT~\cite{wang2020tent}             & 87.57 & 60.63 & 88.57 & 60.29 & 88.07 & 60.46 \\
COME~\cite{zhang2024come}            & 87.28 & 60.14 & 88.43 & 59.83 & 87.86 & 59.99 \\
SAR~\cite{niu2023towards}            & 83.43 & 47.63 & 87.48 & 53.40 & 85.45 & 50.52 \\
MemBN~\cite{kang2024membn}           & 83.51 & 47.58 & 87.58 & 54.32 & 85.55 & 50.95 \\
TEA~\cite{yuan2024tea}               & 87.47 & 60.36 & \underline{88.62} & 60.28 & 88.04 & 60.32 \\
DeYO~\cite{lee2024entropy}           & 87.55 & 60.62 & 88.57 & 60.30 & 88.06 & 60.46 \\
\midrule
SHOT~\cite{liang2020we}              & 87.35 & 60.20 & 88.47 & 59.76 & 87.91 & 59.98 \\
SF(DA)$^2$~\cite{hwang2024sf}        & 87.46 & 60.38 & 88.52 & 60.07 & 87.99 & 60.22 \\
UCon~\cite{xu2025revisiting}         & \underline{87.65} & \underline{60.92} & 88.61 & \underline{60.36} & \underline{88.13} & \underline{60.64} \\
\midrule
\textbf{Ours} & \textbf{88.70} & \textbf{62.57} & \textbf{88.97} & \textbf{62.26} & \textbf{88.84} & \textbf{62.42} \\
\bottomrule
\end{tabular}
\end{table}

\begin{table}[!t]
\centering
\caption{Online TTA results under continual and independent settings. \textbf{Bold}: best, \underline{underlined}: second best within each setting. All values are reported in percent (\%).}
\label{tab:online_tta}
\setlength{\tabcolsep}{5pt}
\begin{tabular}{l|cc|cc|cc}
\toprule
\multirow{2}{*}{\textbf{Method}} &
\multicolumn{2}{c|}{\textbf{CPSC2018}} &
\multicolumn{2}{c|}{\textbf{Georgia}} &
\multicolumn{2}{c}{\textbf{Avg}} \\
& \textbf{AUC} & \textbf{F1} & \textbf{AUC} & \textbf{F1} & \textbf{AUC} & \textbf{F1} \\
\midrule
\textbf{Source Only} & 87.62 & 60.84 & 88.59 & 60.35 & 88.11 & 60.59 \\
\midrule
\multicolumn{7}{c}{\textit{Continual Online TTA}} \\
\midrule
AdaBN~\cite{li2016revisiting}   & 86.11 & 58.81 & 87.94 & \textbf{61.93} & 87.03 & 60.37 \\
TENT~\cite{wang2020tent}        & 84.92 & 53.68 & 87.40 & 56.15 & 86.16 & 54.92 \\
COME~\cite{zhang2024come}       & 87.29 & 60.63 & 88.54 & 60.33 & 87.92 & 60.48 \\
SAR~\cite{niu2023towards}       & \underline{87.63} & \underline{60.86} & \textbf{88.60} & 60.37 & \underline{88.12} & \underline{60.62} \\
MemBN~\cite{kang2024membn}      & 87.43 & 60.44 & 88.38 & 60.65 & 87.91 & 60.55 \\
TEA~\cite{yuan2024tea}          & 87.52 & 60.77 & 88.59 & 60.25 & 88.06 & 60.51 \\
DeYO~\cite{lee2024entropy}      & 87.38 & 60.55 & 88.54 & 60.24 & 87.96 & 60.39 \\
\textbf{Ours}                   & \textbf{88.98} & \textbf{62.53} & \underline{88.55} & \underline{60.94} & \textbf{88.77} & \textbf{61.74} \\
\midrule
\multicolumn{7}{c}{\textit{Independent Online TTA}} \\
\midrule
AdaBN~\cite{li2016revisiting}   & 87.25 & 59.37 & 88.44 & 58.75 & 87.85 & 59.06 \\
TENT~\cite{wang2020tent}        & 87.20 & 60.37 & 88.43 & 59.87 & 87.82 & 60.12 \\
COME~\cite{zhang2024come}       & 87.35 & 60.47 & 88.50 & 60.19 & 87.93 & 60.33 \\
SAR~\cite{niu2023towards}       & \underline{87.64} & \underline{60.90} & \underline{88.61} & \underline{60.41} & \underline{88.13} & \underline{60.66} \\
MemBN~\cite{kang2024membn}      & 85.17 & 45.42 & 85.55 & 46.47 & 85.36 & 45.95 \\
TEA~\cite{yuan2024tea}          & 87.55 & 60.68 & 88.59 & 60.23 & 88.07 & 60.45 \\
DeYO~\cite{lee2024entropy}      & 87.35 & 60.39 & 88.47 & 59.96 & 87.91 & 60.18 \\
\textbf{Ours}                   & \textbf{88.27} & \textbf{62.97} & \textbf{88.63} & \textbf{62.30} & \textbf{88.45} & \textbf{62.63} \\
\bottomrule
\end{tabular}
\end{table}

\begin{table}[!t]
\centering
\caption{Ablation study under offline setting. All values are reported in percent (\%).}
\label{tab:ablation}
\setlength{\tabcolsep}{4.5pt}
\begin{tabular}{ccc|cc|cc|cc}
\toprule
\textbf{SQI} &
\multicolumn{2}{c|}{\textbf{Consistency}} &
\multicolumn{2}{c|}{\textbf{CPSC2018}} &
\multicolumn{2}{c|}{\textbf{Georgia}} &
\multicolumn{2}{c}{\textbf{Avg}} \\
& \textbf{Rhythm} & \textbf{Beat} &
\textbf{AUC} & \textbf{F1} & \textbf{AUC} & \textbf{F1} & \textbf{AUC} & \textbf{F1} \\
\midrule
& & & 87.62 & 60.84 & 88.59 & 60.35 & 88.11 & 60.59 \\
\cmark & & & 88.45 & 61.73 & 88.89 & 61.58 & 88.67 & 61.66 \\
& \cmark & \cmark & 88.55 & 61.95 & 88.91 & 61.82 & 88.73 & 61.89 \\
\cmark & \cmark & & 88.65 & 62.38 & 88.95 & 62.13 & 88.80 & 62.26 \\
\cmark & \cmark & \cmark & \textbf{88.70} & \textbf{62.57} & \textbf{88.97} & \textbf{62.26} & \textbf{88.84} & \textbf{62.42} \\
\bottomrule
\end{tabular}
\end{table}

\textbf{Ablation Study.}
Table~\ref{tab:ablation} validates the necessity of each proposed module under offline TTA.
Removing SQI gating degrades performance, confirming that reliability-aware filtering prevents noisy segments from driving harmful adaptation.
Removing the consistency regularization further reduces Macro-F1 and Macro-AUC, showing that semantic-preserving constraints offer complementary benefits beyond gating.
Finally, beat-level consistency yields a clear additional gain, supporting the role of beat-aware alignment in ECG adaptation.

\section{Conclusion}

We presented BeatRhythm-TTA, an ECG-tailored framework combining SQI-gated pseudo-labeling with dual-level consistency regularization. Across two target domains, our method consistently outperformed representative TTA and SFDA baselines under offline, continual, and independent online protocols. The results highlight that reliability-aware updates and semantically grounded beat-rhythm constraints are critical for stable ECG adaptation under real-world shifts.

\bibliographystyle{splncs04}
\bibliography{refs}

\end{document}